\documentclass[conference]{IEEEtran}

\usepackage{cite}
\usepackage{amsmath,amssymb,amsfonts}
\usepackage{graphicx}
\usepackage{textcomp}
\usepackage{xcolor}
\usepackage[utf8]{inputenc} 
\usepackage[T1]{fontenc}    
\usepackage{hyperref} 
\usepackage{url}            
\usepackage{booktabs}       
\usepackage{amsfonts}       
\usepackage{amsmath}
\usepackage{nicefrac}       
\usepackage{microtype}      
\usepackage{lipsum}		
\usepackage{graphicx}
\usepackage{doi}
\usepackage{algorithm}
\usepackage{algpseudocode}
\usepackage{wrapfig}
\usepackage{multirow}
\usepackage{caption}
\usepackage{footnote}
\usepackage{placeins}
\usepackage{float}

\def\BibTeX{{\rm B\kern-.05em{\sc i\kern-.025em b}\kern-.08em
    T\kern-.1667em\lower.7ex\hbox{E}\kern-.125emX}}

\begin{document}

\title{The Benefits of a Concise Chain of Thought on Problem-Solving in Large Language Models}

\author{\IEEEauthorblockN{1\textsuperscript{st} Matthew Renze}
\IEEEauthorblockA{\textit{Johns Hopkins University} \\
Baltimore, MD, USA \\
mrenze1@jhu.edu}
\and
\IEEEauthorblockN{2\textsuperscript{nd} Erhan Guven}
\IEEEauthorblockA{\textit{Johns Hopkins University} \\
Baltimore, MD, USA \\
eguven2@jhu.edu}
}

\maketitle

\begin{abstract}
In this paper, we introduce Concise Chain-of-Thought (CCoT) prompting. We compared standard CoT and CCoT prompts to see how conciseness impacts response length and correct-answer accuracy. We evaluated this using GPT-3.5 and GPT-4 with a multiple-choice question-and-answer (MCQA) benchmark. CCoT reduced average response length by 48.70\% for both GPT-3.5 and GPT-4 while having a negligible impact on problem-solving performance. However, on math problems, GPT-3.5 with CCoT incurred a performance penalty of 27.69\%. Overall, CCoT leads to an average per-token cost reduction of 22.67\%. All code, data, and supplemental materials are available on GitHub at \href{https://github.com/matthewrenze/jhu-concise-cot}{https://github.com/matthewrenze/jhu-concise-cot}
\end{abstract}


\section{Introduction}

\subsection{Background}

In recent years, Large Language Models (LLMs) have transformed the field of artificial intelligence by offering unprecedented new capabilities for AI systems. As a result, LLMs have become a standard component in many AI systems that automate solutions to real-world problems.

However, to create effective LLM solutions, prompt engineering is often necessary. So, AI systems engineers have developed various techniques to improve the performance of LLMs for specific use cases and problem domains. These techniques include Chain of Thought (CoT) prompting \cite{Mialon2023,White2023}.

\subsection{Chain-of-Thought (CoT) Prompting}
CoT is a prompt engineering technique that instructs the LLM to reason through a problem in a step-by-step manner. Reasoning step-by-step increases the likelihood of the LLM producing a correct solution to the problem. As a result, CoT improves LLM performance on many problem-solving tasks \cite{Kojima2022,Wei2022,Zhou2023}.

There are multiple versions of CoT prompting with various pros and cons. Zero-shot CoT instructs the LLM to “think step-by-step” through the problem in the system prompt \cite{Kojima2022,Zhou2023}. Few-shot CoT provides a series of examples as problem-solution pairs with the CoT explicitly stated in each example solution \cite{Wei2022}.

CoT prompting has been shown to improve LLM performance by up to 80\% for certain problem tasks and problem domains \cite{Kojima2022}. However, this performance increase comes at the expense of increased LLM response length. As a result, the cost of using the LLM with CoT grows in proportion to response length.

\subsection{Concise Prompting}
Concise prompting is a prompt-engineering technique used to reduce LLM response verbosity. The main benefit is that it decreases the per-token cost of using the LLM. In addition, it can reduce the LLM’s energy consumption, minimize response time, and improve communication efficiency with the end user.

There are two main implementations of concise prompting. Zero-shot prompting instructs the LLM to “be concise” in its response \cite{Crispino2023,Kadous2023}. Few-shot prompting requires the prompt engineer to create a series of problem-solution example pairs with concisely written text in each example solution.

While concise prompting is beneficial for reducing resource costs, it may negatively impact the performance of the LLM on some problem-solving tasks \cite{Crispino2023}. This is because the LLM requires additional verbosity to fully elaborate the steps in its thought process to produce a correct solution.

As a result, prompt engineers often provide LLMs with instructions and examples designed to ensure higher verbosity so that all steps in the LLM’s thought process are explicitly stated. However, the adoption of verbose CoT conventions appears to be based on anecdotal rather than empirical evidence.

\subsection{Concise Chain-of-Thought (CCoT)}
It is still an open question how conciseness impacts the problem-solving capabilities and response length of an LLM with CoT. To answer this question, we combined concise prompting and CoT to create Concise Chain-of-Thought (CCoT) prompting. 

CCoT prompting is a novel prompt engineering technique that combines the effectiveness of CoT prompting with the efficiency of concise prompting. It attempts to produce a chain of thought that leads to a correct solution with the shortest possible response length.

CCoT is achieved by instructing the LLM to both "think step-by-step" and "be concise". In addition, the LLM is provided with few-shot examples that include a sample problem and a concise solution rather than a standard (i.e., verbose) solution.

For an example of the CCoT prompt and few-shot examples, please see Figure \ref{fig:concise-cot-prompt} in the appendix.

\subsection{Contribution}
Our research builds upon prior literature by determining how concise prompting impacts CoT prompting. These results have practical implications for AI engineers building software solutions with LLMs.

First, if CCoT reduces response length, then AI systems engineers can use CCoT to reduce LLM costs. Third-party LLM APIs are typically priced per token \cite{Anyscale2023, OpenAI2023_pricing}. So, reducing response length will reduce total costs. Reducing response length also reduces energy consumption and response times.

Second, if CCoT does not decrease performance, then there is no performance penalty for implementing CCoT. As a result, AI systems engineers should prefer CCoT over standard CoT.

Finally, these results have theoretical implications for AI researchers studying CoT reasoning in LLMs. If we can reduce the length of a CoT without impacting performance, then only some aspects of a CoT are relevant to the LLM’s problem-solving performance. This discovery raises new questions about which specific tokens or aspects of an LLM’s CoT are necessary vs. which are superfluous.

\subsection{Prior Literature}
Few-shot CoT prompting was introduced by Wei et al. in January 2022 \cite{Wei2022}. Zero-shot CoT was introduced four months later by Kojima et al. \cite{Kojima2022}. The zero-shot CoT method was then further refined by Zhou et al. using the Automatic Prompt Engineer (APE) method in October 2022 \cite{Zhou2023}.

In September 2022, Madaan and Yazdanbakhsh attempted to separate symbols, patterns, and text to determine their individual effects on CoT and develop their own concise CoT prompt technique \cite{Madaan2022a}. However, as of January 2023, their paper was withdrawn from publication due to technical issues \cite{Madaan2022b}.

Concise prompting, in general, has not been studied to the same extent as CoT prompting. It has been used in scientific research papers for practical reasons but has not been studied directly. Most guidance on concise prompting comes from best practices in the prompt-engineering community.

After an extensive literature search, we found no research papers specifically studying concise prompting. In addition, we could not find any other publications exploring concise CoT prompting.

\section{Methods}
\subsection{Data}

Our test dataset consists of Multiple-Choice Question-and-Answer (MCQA) problems from standard LLM benchmarks. 

We reviewed existing literature to identify a set of candidate benchmarks that spanned multiple problem domains and difficulty levels. The ten source benchmarks were selected from ARC, AGIEval, HellaSwag, and MedMCQA \cite{Clark2018,Zellers2019,Pal2022,Zhong2023,Liu2020,Wang2021}.

We preprocessed the data into a standard data format and randomly selected 100 questions from each of the ten benchmarks to create an exam with 1,000 MCQA problems.

For a complete list of the source problem sets used to create the MCQA test set, see Table \ref{tab:exams} in the appendix. For a sample MCQA problem, see Figure \ref{fig:sample-mcqa-problem} in the appendix.

\subsection{Models}
We used two popular LLMs in our experiment -- namely GPT-3.5 and GPT-4. 

GPT-3.5 is a Generative Pre-trained Transformer (GPT) created by OpenAI and publicly released as “ChatGPT” in November 2022 \cite{Brown2020,OpenAI2022}. GPT-4 is a more powerful GPT with more advanced capabilities \cite{OpenAI2023,OpenAI2023_gpt4}. However, GPT-4 costs roughly 30x more per output token than GPT-3.5 \cite{OpenAI2023_pricing}.

GPT-3.5 and GPT-4 can be accessed via an application programming interface (API) hosted by OpenAI or Microsoft. We used Microsoft Azure OpenAI Service for our experiments -- though our results should hold for either platform -- since they both use the same underlying foundational models \cite{Microsoft2023,OpenAI2023_models,OpenAI2023_api}.

\subsection{Prompts}
We used three prompt-engineering techniques in our experiment. These prompts consisted of an answer-only prompt, a standard (i.e., verbose) CoT prompt, and a concise CoT prompt.

The answer-only prompt instructed the LLM to respond with only the answer to the question. The prompt included a single (i.e., one-shot) example to demonstrate how to complete the task. This prompt provided a baseline for evaluating the minimum response length and task performance.

The standard CoT prompt instructed the LLM to think step-by-step through its thought process before answering. The one-shot example included a verbose CoT for each reasoning step in the solution. This prompt provided an upper bound for response length and task performance. 

The CCoT prompt included the standard CoT prompt but also had a final instruction to “Be concise.” The one-shot example also included a more concise CoT in its solution. This prompt allowed us to compare CCoT to both the answer-only and standard CoT prompts.

For samples of all three prompts, see Figures \ref{fig:answer-only-prompt}, \ref{fig:verbose-cot-prompt}, and \ref{fig:concise-cot-prompt} in the appendix.

\subsection{Metrics}
To compare the performance of our prompts, we measured the LLM’s response length and correct-answer accuracy.

Response length was measured as the number of output tokens produced by the LLM in its response. A token is the smallest unit of information an LLM can process. Both GPT-3.5 and GPT-4 work with sub-word tokens. There are approximately 1.3 tokens per word on average \cite{OpenAI2023_tokens}.

Performance was measured as the number of correctly answered questions divided by the total number of questions asked. This performance metric allowed us to directly compare the LLM’s ability to correctly solve a problem presented in the MCQA format.

\subsection{Analysis}
To determine the statistical significance of our results, we performed a set of Mann-Whitney U (MWU) tests \cite{Mann1947,SciPy2023}. 

For the response-length results, we performed an MWU test on the distributions of the CoT and CCoT response lengths. For the performance results, we performed an MWU test on the distributions of the correct-answer accuracy for standard CoT and CCoT.

We used an MWU test instead of a t-test due to the non-normal distribution of the data. A Shapiro-Wilk normality test indicated that the response-length data and accuracy data were not normally distributed. This condition violates the normality assumption of the t-test, so an MWU test was used instead.

\section{Results}

\begin{figure}[h!]
    \centering
    \includegraphics[width=\linewidth]{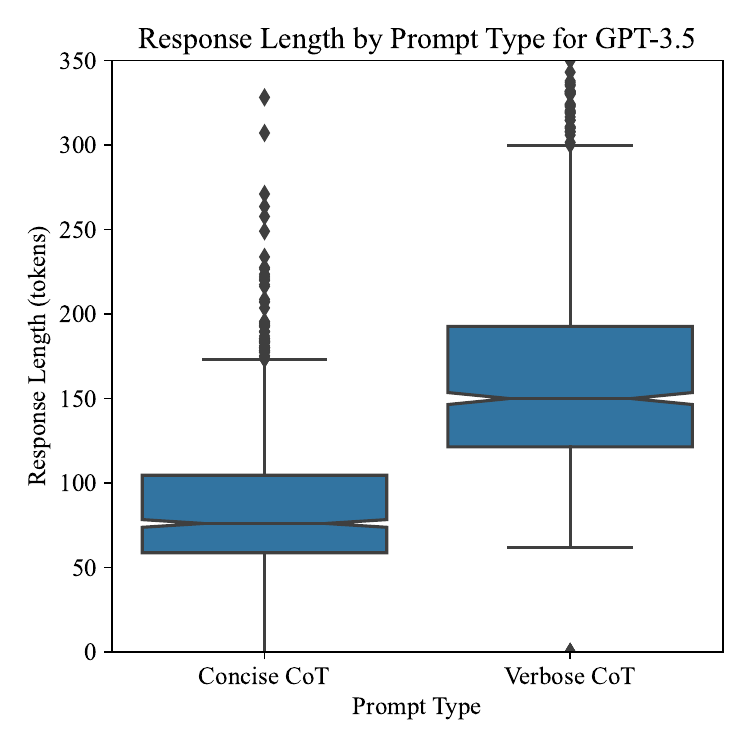}
    \caption{GPT-3.5 with CCoT reduces response length by an average of 47.62\% compared to verbose CoT.}
    \label{fig:response-length-by-prompt-for-gpt-3.5}

    \includegraphics[width=\linewidth]{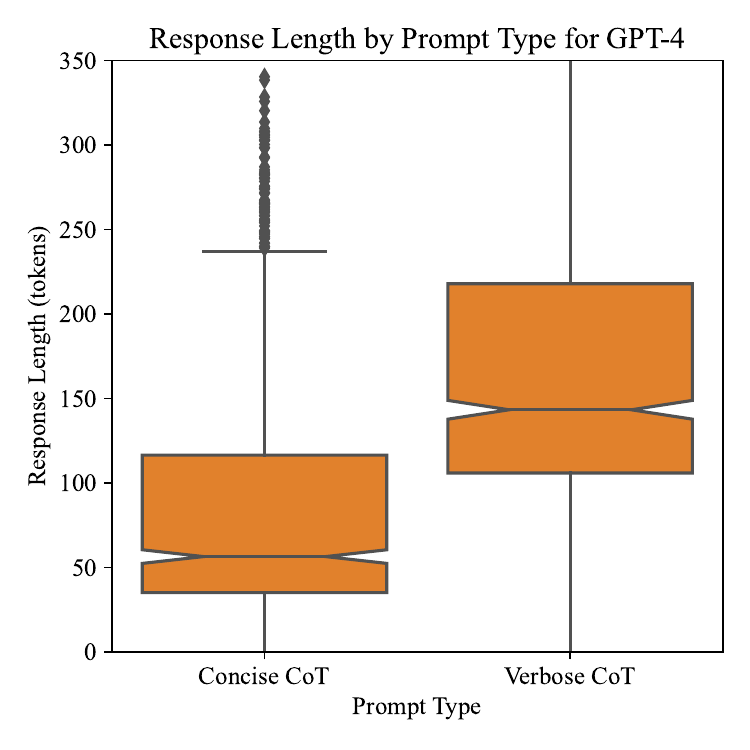}
    \caption{GPT-4 with CCoT reduces response length by an average of 49.77\% compared to verbose CoT.}
    \label{fig:response-length-by-prompt-for-gpt-4}
\end{figure}

\subsection{Response Length}
Our analysis revealed that CCoT reduced average response length by 48.70\% compared to CoT. 

For GPT-3.5, we saw a 47.62\% decrease in average response length when comparing CCoT to CoT. The MWU test yielded $(U = 873,046.00, p < 0.001)$, indicating a significant difference between the two distributions. 

For GPT-4, we saw a 49.77\% decrease in average response length between CCoT and CoT. The MWU test yielded $(U = 807,523.50, p < 0.001)$, also indicating a significant difference between the distributions.

A visual analysis of the data supports these findings (see Figures~\ref{fig:response-length-by-prompt-for-gpt-3.5} and~\ref{fig:response-length-by-prompt-for-gpt-4}).

A further analysis of response length by exam shows this decrease in response length holds across all problem domains. This pattern can be observed for both GPT-3.5 and GPT-4 (see Figures~\ref{fig:response-length-by-prompt-and-exam-for-gpt-3.5} and~\ref{fig:response-length-by-prompt-and-exam-for-gpt-4} in the appendix).

\subsection{Performance}

\begin{figure}[h!]
    \centering
    \includegraphics[width=\linewidth]{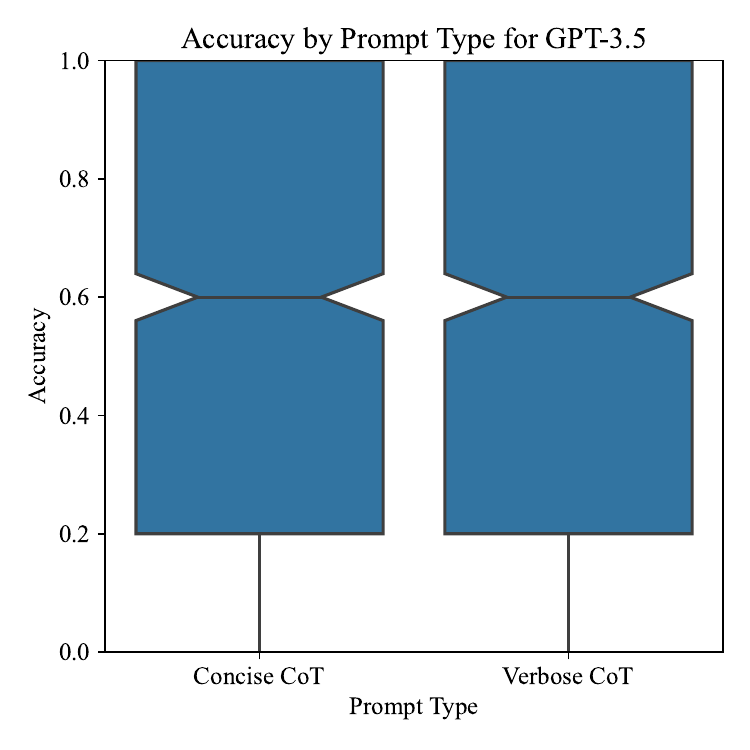}
    \caption{GPT-3.5 with CCoT has similar accuracy compared to verbose CoT.}
    \label{fig:accuracy-by-prompt-for-gpt-3.5}

    \includegraphics[width=\linewidth]{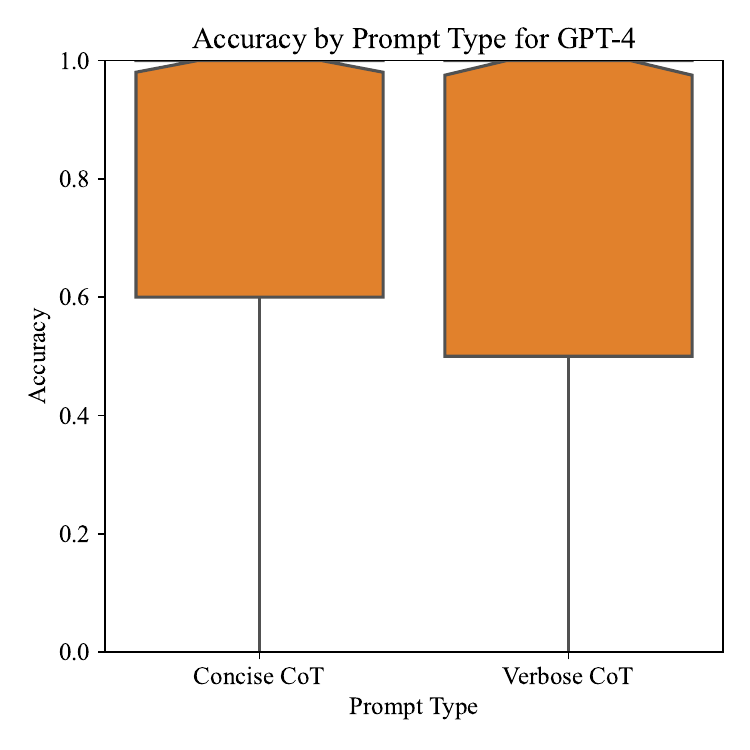}
    \caption{GPT-4 with CCoT has similar accuracy compared to verbose CoT. (Note: the median value for both prompts is 1.0, making the box plot more difficult to interpret visually.)}
    \label{fig:accuracy-by-prompt-for-gpt-4}
\end{figure}

Our analysis revealed that CCoT did not reduce problem-solving performance in any statistically meaningful way compared to standard CoT.

For GPT-3.5, we saw a 2.95\% decrease in average accuracy when comparing CCoT to CoT. The MWU test yielded $(U = 507,648.50, p = 0.55)$, indicating the difference between the distributions was not statistically significant.

For GPT-4, we saw an increase in average accuracy of 0.25\% when comparing CCoT to CoT prompting. The MWU test yielded $(U = 485,396.00, p = 0.21)$, also indicating a non-significant difference between the distributions.

A visual analysis of these data supports these findings (see Figures~\ref{fig:accuracy-by-prompt-for-gpt-3.5} and~\ref{fig:accuracy-by-prompt-for-gpt-4}).

A more in-depth analysis of accuracy by exam revealed that GPT-3.5 with CCoT produced a statistically significant decrease in performance \textit{only} for math problems (i.e., AQUA-RAT and SAT Math) compared to CoT. Other problem domains saw no decrease in accuracy (see Figure \ref{fig:accuracy-by-prompt-and-exam-for-gpt-3.5} in the appendix).

For these two math exams, GPT-3.5 with CCoT resulted in an average reduction in accuracy of 27.69\%. An MWU test on the distributions of CoT and CCoT yielded $(U = 26,546.00, p < 0.001)$. Thus, for GPT-3.5 \textit{only}, math problems appear to be a special case where conciseness negatively impacts performance.

On the other hand, GPT-4 with CCoT did not result in a statistically significant decrease in performance on math problems compared to CoT. Both CCoT and CoT had roughly equivalent performance on all problem domains. CCoT also significantly outperformed the answer-only prompt on math problems (see Figure \ref{fig:accuracy-by-prompt-and-exam-for-gpt-4} in the appendix).

\subsection{Cost Analysis}

To understand the practical implications of these results, we computed the cost of solving all 1,000 problems using the current per-token pricing model for GPT-3.5 and GPT-4. \footnote{For our study, we solved each MCQA problem 10 times. So, our total cost was roughly 10x the values we are reporting.} 

Currently, GPT-3.5 is priced at \$0.001 per 1,000 input tokens and \$0.002 per 1,000 output tokens. \footnote{All prices are in US Dollars (USD).} GPT-4 is priced at \$0.03 per 1,000 input tokens and \$0.06 per 1,000 output tokens \cite{OpenAI2023_pricing}. As a result, CCoT produced a total cost savings of 21.85\% for GPT-3.5 and 23.49\% for GPT-4 (see Table~\ref{tab:cost-analysis}). These cost savings should scale linearly.

\begin{table}[!h]
\centering
\begin{tabular}{lrrrr}
\toprule
 & GPT-3.5 & GPT-3.5 & GPT-4 & GPT-4 \\
 & CoT & CCoT & CoT & CCoT \\
\midrule
Input Cost (\$) & 0.55 & 0.51 & 16.37 & 15.29 \\
Output Cost (\$) & 0.33 & 0.17 & 10.53 & 5.29 \\
Total Cost (\$) & 0.88 & 0.69 & 26.90 & 20.58 \\
Cost Savings (\%) & & \textit{21.85} & & \textit{23.49} \\
\bottomrule
\end{tabular}
\caption{CCoT significantly reduces total costs per 1,000 problems for both GPT-3.5 and GPT-4.}
\label{tab:cost-analysis}
\end{table}

\section{Discussion}
\subsection{Limitations}
There are several limitations in this research study:

First, our study involved only two LLMs – both of which are versions of OpenAI's GPT. As a result, these findings may not generalize to other proprietary and open-source LLMs like Llama 2, PaLM, and Claude.

Second, our study only tested a single type of CoT and CCoT prompt. As a result, other variations of CoT and CCoT prompts may produce different results.

Third, our study was limited to ten problem domains. As a result, these findings may not hold for other problem domains or sub-domains within each problem domain.

Finally, in the case of GPT-4, the median accuracy was 1.0. As a result, the data were compressed at the ceiling of accuracy. This compression may have caused issues with our statistical analysis of performance due to a ceiling effect.

\subsection{Implications}
These results have practical implications for AI engineers building LLM-based software solutions. 

Since most proprietary LLM APIs charge on a per-token pricing model, reducing the number of output tokens in an LLM’s response yields direct cost savings \cite{Anyscale2023,OpenAI2023_pricing}. In addition, obtaining more concise responses without sacrificing accuracy leads to direct savings in energy consumption and response times. 

These results also have theoretical implications for AI researchers studying how LLMs perform step-by-step reasoning using CoT. 

CCoT can reduce the number of output tokens by roughly half while maintaining the same level of accuracy. As a result, only a subset of the CoT tokens contribute to the performance of the LLM. This opens new questions about which
aspects of a CoT lead to a correct solution, and which are superfluous. 

\subsection{Future Research}
To improve upon this research, we suggest additional follow-up experiments:

First, we recommend performing additional experiments with other proprietary and open-source LLMs. It would be beneficial to know if these results hold for all LLMs or only GPT-3.5 and GPT-4.

Second, we recommend testing more CCoT prompt variations. Instructing the LLM to be even more concise in the system prompt may further decrease response length. In addition, more concisely written few-shot examples may produce even more succinct responses without sacrificing accuracy.

Third, we recommend testing more task types and problem domains beyond those tested in our MCQA test set. These results may not generalize to non-MCQA task types. In addition, they may not generalize to other problem domains. Other math sub-domains may also exhibit behaviors different from what we observed.

Finally, we recommend performing a more in-depth error analysis to improve our understanding of the nature of CCoT errors. The LLMs may be making specific types of errors in their CoT that impact their performance. Understanding these errors may provide additional insight into how to mitigate them.

\section{Conclusion}
In this research study, we introduced the CCoT prompt-engineering technique. 

We demonstrated that CCoT prompting reduced response token length for GPTs by 48.70\% while maintaining comparable performance to standard CoT. GPT-4 incurred no performance penalty in any problem domain. However, for math problems, GPT-3.5 incurred a 27.69\% reduction in accuracy.

In practice, CCoT can reduce the per-token cost of solving multi-step problems by 22.67\%. These cost savings also extend to reduced energy consumption and shorter LLM response times.

\bibliographystyle{IEEEtran}
\bibliography{references}

\newpage
\onecolumn
\appendix

\subsection{Response Length for GPT-3.5}
\vspace{-1.0em}

\begin{figure*}[h!]
    \centering
    \includegraphics[width=0.95\textwidth]{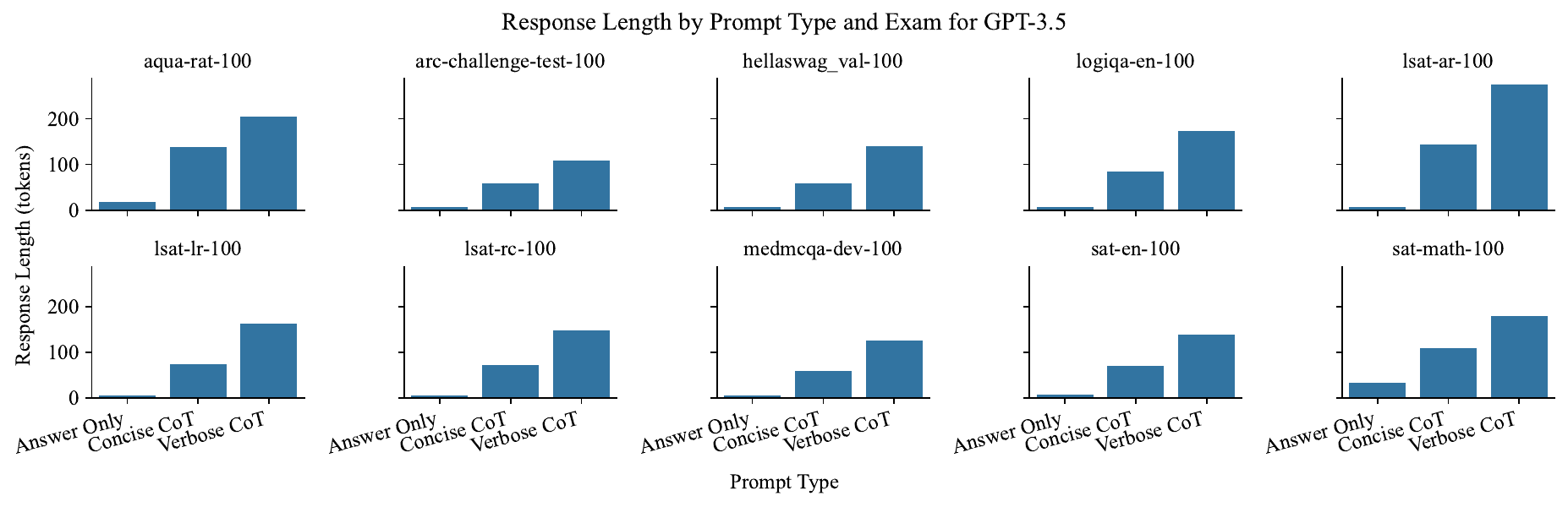}
    \caption{GPT-3.5 with CCoT decreases response length significantly across all problem domains compared to CoT.}
    \label{fig:response-length-by-prompt-and-exam-for-gpt-3.5}
\end{figure*}

\subsection{Response Length for GPT-4}
\vspace{-1.0em}

\begin{figure*}[h!]
    \centering
    \includegraphics[width=0.95\textwidth]{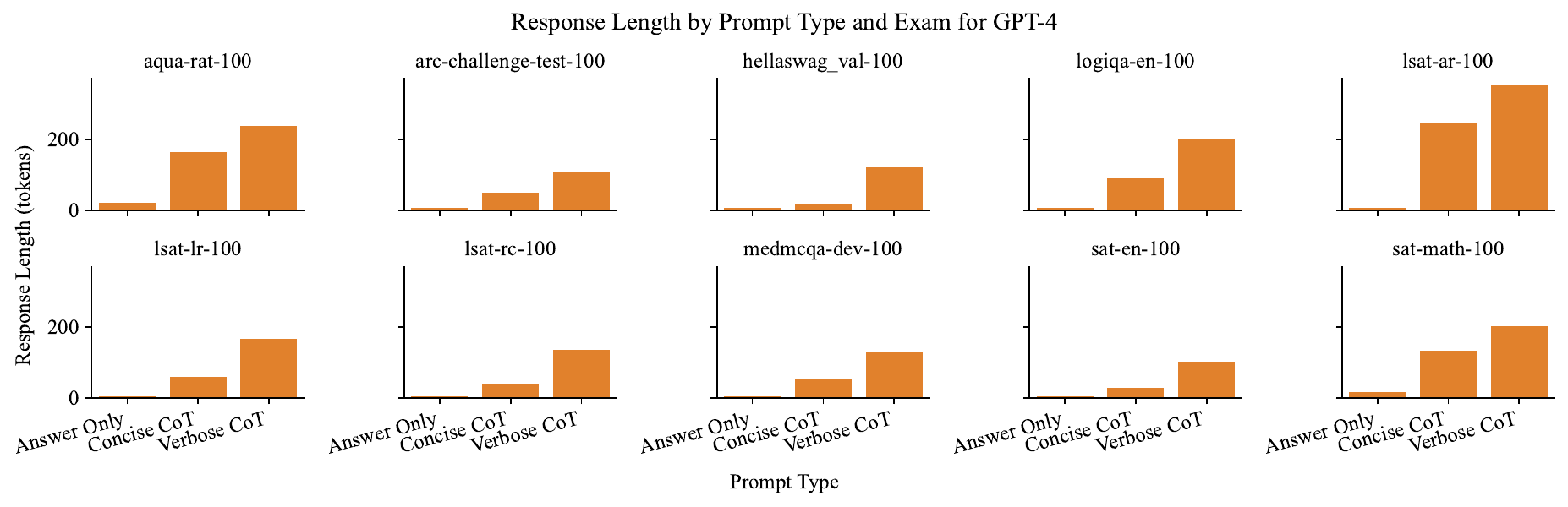}
    \caption{GPT-4 with CCoT decreases response length significantly across all problem domains compared to CoT.}
    \label{fig:response-length-by-prompt-and-exam-for-gpt-4}
\end{figure*}

\subsection{Performance for GPT-3.5}
\vspace{-1.0em}

\begin{figure*}[h!]
    \centering
    \includegraphics[width=0.95\textwidth]{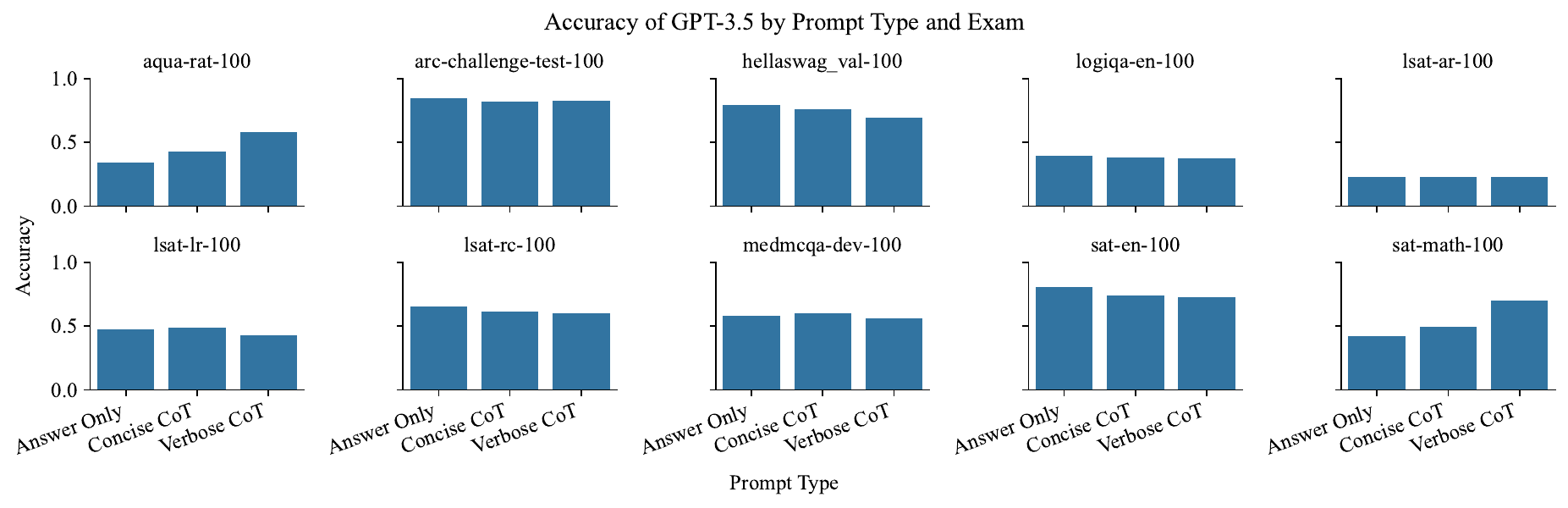}
    \caption{Verbose CoT prompting with GPT-3.5 significantly improved performance on math problems (i.e., AQUA-RAT and SAT Math) compared to CCoT but had minimal impact on other problem domains.}
    \label{fig:accuracy-by-prompt-and-exam-for-gpt-3.5}
\end{figure*}

\newpage

\subsection{Performance for GPT-4}
\vspace{-1.0em}

\begin{figure*}[h!]
    \centering
    \includegraphics[width=0.95\textwidth]{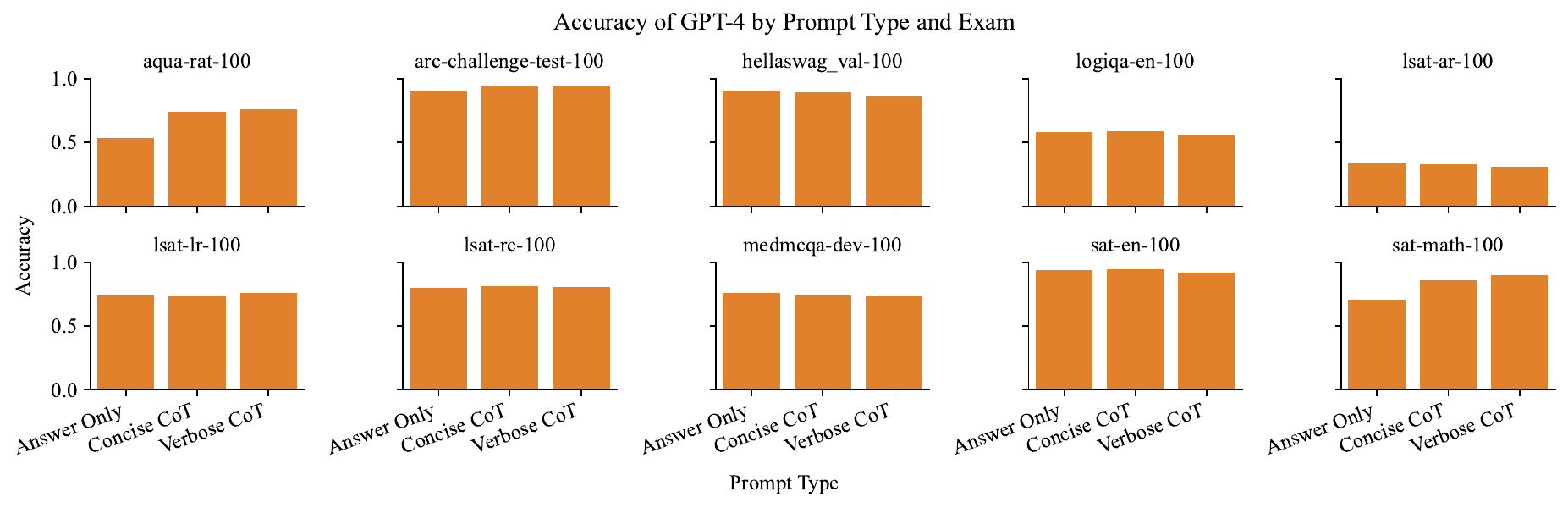}
    \caption{Verbose CoT and CCoT prompting with GPT-4 significantly improved accuracy for math problems (i.e., AQUA-RAT and SAT Math) but had minimal impact on other problem domains.
}
    \label{fig:accuracy-by-prompt-and-exam-for-gpt-4}
\end{figure*}

\subsection{Problem Sets}

\begin{table*}[ht]
    \caption{Problem sets used to create the 1,000-question multi-domain MCQA exam.}
    \label{tab:exams}
    \vskip 0.15in
    \begin{center}
        \begin{tabular}{llllrl}
            \toprule
            Problem Set & Benchmark & Domain & Questions & License & Source \\
            \midrule
            ARC Challenge Test & ARC & Science & 1,173 & CC BY-SA & \cite{Clark2018} \\
            AQUA-RAT & AGI Eval & Math & 254 & Apache v2.0 & \cite{Zhong2023} \\
            Hellaswag Val & Hellaswag & Common Sense Reasoning & 10,042 & MIT & \cite{Zellers2019} \\
            LogiQA (English) & AGI Eval & Logic & 651 & GitHub & \cite{Zhong2023, Liu2020} \\
            LSAT-AR & AGI Eval & Law (Analytic Reasoning) & 230 & MIT & \cite{Zhong2023, Wang2021} \\
            LSAT-LR & AGI Eval & Law (Logical Reasoning) & 510 & MIT & \cite{Zhong2023, Wang2021} \\
            LSAT-RC & AGI Eval & Law (Reading Comprehension) & 260 & MIT & \cite{Zhong2023, Wang2021} \\
            MedMCQA Valid & MedMCQA & Medicine & 6,150 & MIT & \cite{Pal2022} \\
            SAT-English & AGI Eval & English & 206 & MIT & \cite{Zhong2023} \\
            SAT-Math & AGI Eval & Math & 220 & MIT & \cite{Zhong2023} \\
            \bottomrule
        \end{tabular}
    \end{center}
    \vskip -0.1in
\end{table*}
{\footnotesize Note: The GitHub repository for LogiQA does not include a license file. However, the paper and readme.md file states that "The dataset is freely available."}

\subsection{Sample MCQA Problem}
\begin{figure}[h]
\centering
\scriptsize
\begin{verbatim}
{
  "id": 3, 
  "source": "agi-eval/aqua-rat", 
  "source_id": 35, 
  "topic": "Math", 
  "context": "", 
  "question": "A rectangular solid, 3 x 4 x 15, is inscribed in a sphere, 
               so that all eight of its vertices are on the sphere. 
               What is the diameter of the sphere?", 
  "choices": {
    "A": " 13.3542", 
    "B": " 15.8113", 
    "C": " 18.3451", 
    "D": " 19.5667", 
    "E": " 20.8888"}, 
  "answer": "B", 
  "solution": "In an inscribed rectangle in a sphere, we will have a line joining 
               the opposite vertices as the diameter. According to the Pythagoras theorem,
               sides 3, 4 give diagonal as 5 ==> with 5 and 15, we get 5sqrt(10).
               5sqrt(10) or 15.8113 is the diameter of the sphere.\nanswer = B"
}
\end{verbatim}
\caption{Sample of a multiple-choice question in standardized data format – with whitespace added for readability.}
\label{fig:sample-mcqa-problem}
\end{figure}

\section{Prompts}

\newpage
\subsection{Answer-Only Prompt}
\vspace{-1.0em}

\begin{figure}[h!]
    \centering
    \scriptsize
    \captionsetup{skip=-10pt}
    \begin{verbatim}
[System Prompt]
You are an intelligent assistant.
Your task is to answer the following multiple-choice questions.
You MUST answer the question using the following format 'Action: Answer("[choice]")'  
The parameter [choice] is the letter or number of the answer you want to select (e.g. "A", "B", "C", or "D").
For example, 'Answer("C")' will select choice "C" as the best answer.
You MUST select one of the available choices; the answer CANNOT be "None of the Above".

[Example Problem]
Question: What is the capital of the state where Johns Hopkins University is located?
Choices:
  A: Baltimore
  B: Annapolis
  C: Des Moines
  D: Las Vegas

[Example Solution]
Action: Answer("B")

    \end{verbatim}
    \caption{Sample of the answer-only system prompt and one-shot example}
    \label{fig:answer-only-prompt}
\end{figure}

\subsection{Verbose CoT Prompt}
\vspace{-1.0em}

\begin{figure}[h!]
    \centering
    \scriptsize
    \captionsetup{skip=-10pt}
    \begin{verbatim}
[System Prompt]
You are an intelligent assistant.
Your task is to answer the following multiple-choice questions.
Think step-by-step through the problem to ensure you have the correct answer.
Then, answer the question using the following format 'Action: Answer("[choice]")'
The parameter [choice] is the letter or number of the answer you want to select (e.g. "A", "B", "C", or "D").
For example, 'Answer("C")' will select choice "C" as the best answer.
You MUST select one of the available choices; the answer CANNOT be "None of the Above".

[Example Problem]
Question: What is the capital of the state where Johns Hopkins University is located?
Choices:
A: Baltimore
B: Annapolis
C: Des Moines
D: Las Vegas

[Example Solution]
Thought:
Johns Hopkins University is located in Baltimore.
Baltimore is a city located in the State of Maryland.
The capital of Maryland is Annapolis.
Therefore, the capital of the state where Johns Hopkins University is located is Annapolis.
The answer is B:
    \end{verbatim}
    \caption{Sample of standard (i.e., verbose) CoT system prompt and one-shot example}
    \label{fig:verbose-cot-prompt}
\end{figure}

\subsection{Concise CoT Prompt}
\vspace{-1.0em}

\begin{figure}[h!]
    \centering
    \scriptsize
    \captionsetup{skip=-10pt}
    \begin{verbatim}
[System Prompt]
You are an intelligent assistant.
Your task is to answer the following multiple-choice questions.
Think step-by-step through the problem to ensure you have the correct answer.
Then, answer the question using the following format 'Action: Answer("[choice]")'
The parameter [choice] is the letter or number of the answer you want to select (e.g. "A", "B", "C", or "D").
For example, 'Answer("C")' will select choice "C" as the best answer.
You MUST select one of the available choices; the answer CANNOT be "None of the Above".
Be concise.
[Example Problem]
Question: What is the capital of the state where Johns Hopkins University is located?
Choices:
A: Baltimore
B: Annapolis
C: Des Moines
D: Las Vegas
[Example Solution]
Thought:
Johns Hopkins University is located in Baltimore, Maryland.
The capital of Maryland is Annapolis.
Action: Answer("B")
    \end{verbatim}
    \caption{Sample of CCoT system prompt and one-shot example}
    \label{fig:concise-cot-prompt}
\end{figure}

\end{document}